\pdfoutput=1
\documentclass[11pt]{article}

\usepackage{acl}
\usepackage{times}
\usepackage{latexsym}

\usepackage[T1]{fontenc}
\usepackage[utf8]{inputenc}

\usepackage{microtype}

\usepackage{graphicx}
\usepackage{subcaption}
\usepackage{comment}
\usepackage{amsmath}
\usepackage[marginal]{footmisc}

\title{Multimodal Multi-loss Fusion Network for Sentiment Analysis}
\author {
    Zehui Wu\footnotemark[1],\textsuperscript{\rm 1}
    Ziwei Gong\footnotemark[1], \textsuperscript{\rm 1}
    Jaywon Koo, \textsuperscript{\rm 1}
    Julia Hirschberg \textsuperscript{\rm 1}\\
  Department of Computer Science\\Columbia University\\
  \texttt{\{zw2804, zg2272, jk4541\}@columbia.edu} \\
  \texttt{julia@cs.columbia.edu}
}

\begin{document}
\maketitle
\renewcommand{\thefootnote}{\fnsymbol{footnote}} 
\footnotetext[1]{These authors contributed equally to this work.} 
\footnotetext{Code is released at: \url{https://github.com/zehuiwu/MMML}} 

% Jeff: change all the emotion detection into sentiment detection
% Jeff: The abstract and introduction should focus on the feature selection, multi-loss training, and model variation, (context). Add more references for the cross-modality encoder. This fusion network is not novel, but our feature selection, multi-loss training, and the analysis for the model variation are novel. 

\begin{abstract}
This paper investigates the optimal selection and fusion of feature encoders across multiple modalities and combines these in one neural network to improve sentiment detection. We compare different fusion methods and examine the impact of multi-loss training within the multi-modality fusion network, identifying surprisingly important findings relating to subnet performance. We have also found that integrating context significantly enhances model performance. Our best model achieves state-of-the-art performance for three datasets (CMU-MOSI, CMU-MOSEI and CH-SIMS). These results suggest a roadmap toward an optimized feature selection and fusion approach for enhancing sentiment detection in neural networks.

 % We compare different fusion methods, encompassing concatenation and an array of transformer fusion networks. We also examine the impact of multi-loss training within the multi-modality fusion network identifying useful findings relating to subnet performance. 

\end{abstract}

\section{Introduction}

In recent years, the multimodal affective computing field has seen significant advances in feature extraction and multimodal fusion methodologies in recent years \cite{garg-etal-2022-multimodality}, enabling a more nuanced understanding of human emotions by effectively synthesizing audio, text, and visual signals \cite{sun-etal-2023-layer, yu2021le}. This study presents a series of experiments that delve into feature selection, comparative analysis of fusion network performance, multi-loss training, and context modeling utilizing audio and text from three datasets: CMU-MOSI \cite{zadeh2016mosi}, CMU-MOSEI \cite{bagher-zadeh-etal-2018-multimodal} , and CH-SIMS \cite{bagher-zadeh-etal-2018-multimodal}. Our research aims to pioneer state-of-the-art (SOTA) approaches in affective computing tasks, achieved by identifying optimal features for each modality, devising the most effective methods for their fusion, and refining training methodologies to enhance performance.

Hand-crafted feature extraction algorithms often lack flexibility and generalization across diverse tasks. To overcome these limitations, recent studies have proposed fully end-to-end models that jointly optimize feature extraction and learning processes \cite{dai-etal-2021-multimodal, han-etal-2021-improving}. Our work leverages feature representations derived from pre-trained models across different modalities, combining them into an end-to-end framework, which provides a comprehensive and adaptable solution for multimodal affective feature computation. 

In multimodal fusion, the challenge lies in effectively fusing diverse signals, including natural language, facial gestures, and acoustic behaviors. Methods like the Tensor Fusion Network (TFN)\cite{zadeh2017tensor} have been proposed to model intra-modality and inter-modality interactions. More recently, transformer encoder structures with cross-modal attention have gained popularity for integrating multimodal data \cite{tsai2019multimodal, 9765342}, with continous efforts to improve representations of multimodal information \cite{qian-etal-2023-sentiment, hu-etal-2022-unimse}. Building from these ideas, we propose a robust fusion network structure that integrates cross-modal attention and self-attention, with additional feed-forward layers to refine the representations. Furthermore, our approach experiments with restoring original signals during the fusion process.

Our experimental analysis has identified the most efficacious features for different modalities and compared various fusion network methods for amalgamating audio and text signals. Our findings reveal that the incorporation of audio signals consistently elevates performance metrics. More notably, our transformer fusion network demonstrates a remarkable enhancement of results and and achieves state-of-the-art performances across all datasets. Additionally, our exploration into multi-loss training has yielded two significant observations: first, the utilization of distinct labels for each modality in multi-loss training markedly benefits the models' performance; second, training with multimodal features not only boosts overall model performance but also notably enhances accuracy in the single-modality subnet. Furthermore, we compared two context modeling methods and found that contextual integration significantly amplifies model performance across all metrics. These novel findings have advanced our understanding of multi-modal sentiment analysis.

\section{Related Work}
%\subsection{Feature Extraction}
Existing research on multimodal affective computing often employs hand-crafted algorithms to perform initial feature representation extraction and retrieve some fixed representations for each modality \cite{Shenoy_2020, delbrouck-etal-2020-transformer}. However, for these, the extracted features are static and lack the flexibility to be further fine-tuned for different target tasks; also, the manual determination of feature extraction algorithms can lead to sub-optimal performance due to constraints in generalization across diverse tasks \cite{dai-etal-2021-multimodal}. 
To address these issues, recent studies have proposed fully end-to-end models, effectively bridging the gap between feature extraction and learning processes \cite{dai-etal-2021-multimodal, 10.1145/3366423.3380000}. Our research also emphasizes an end-to-end structure that optimizes both phases jointly, presenting a comprehensive and adaptable solution for multimodal affective feature computation.

Lexical features, owing to pre-training on expansive corpora through Transformer-based models, often outperform other modalities. Some recent work aims to improve model performance by incorporating speech information inside the text model such as SPECTRA \cite{yu2023speechtext}, by pre-training a speech-text transformer model to capture the speech-text alignment effectively. A similar innovative method is the Transformer-Based Speech-Prefixed Language Model (TEASEL) \cite{arjmand2021teasel}, which incorporates speech as a dynamic prefix along with the textual.

% Following evolving trends in feature extraction and selection methodologies in multimodal affective computing research, our research seeks to extract speech and text signals using two pre-trained transformer-based models in an end-to-end structure and demonstrates that our model outperforms the other methods.

%\subsection{Multimodal Fusion network}
Many studies have explored multimodal human language time-series data, which typically includes a mixture of natural language, facial gestures, and acoustic behaviors. However, fusing these into a unified representation presents a significant challenge due to the variable sampling rates across modalities and the difficulty in determining intra-modality dependencies. Various methods have been proposed to model the interaction across modalities \cite{barezi-fung-2019-modality}, such as the Tensor Fusion Network \cite{zadeh2017tensor}, which utilizes the Cartesian product of different modalities to model both intra-modality and inter-modality interactions. More recent work has shifted toward employing transformer encoder structures to integrate these signals via cross-modality attention. The MULT model \cite{tsai2019multimodal} has pioneered this approach, introducing directional pairwise cross-modal attention. This method allows for interaction between multimodal sequences across distinct time steps and inherently adapts streams from one modality to another.  
Further research has also leveraged this concept of cross-modality attention \cite{9928357, paraskevopoulos2022mmlatch}, yielding valuable insights into how multimodal data can be processed more effectively. We enhance this approach by employing a self-attention encoder and a feed-forward network to further optimize the multimodal representation after one modality is projected into another using the cross-modality attention module, thus enriching our ability to process and understand multimodal data.
% Sara: we may cut down the last paragraph a bit since it's pretty long

\begin{figure*}[ht]
  \centering
  \begin{minipage}{0.5\textwidth}
    \centering
    \includegraphics[width=1\textwidth]{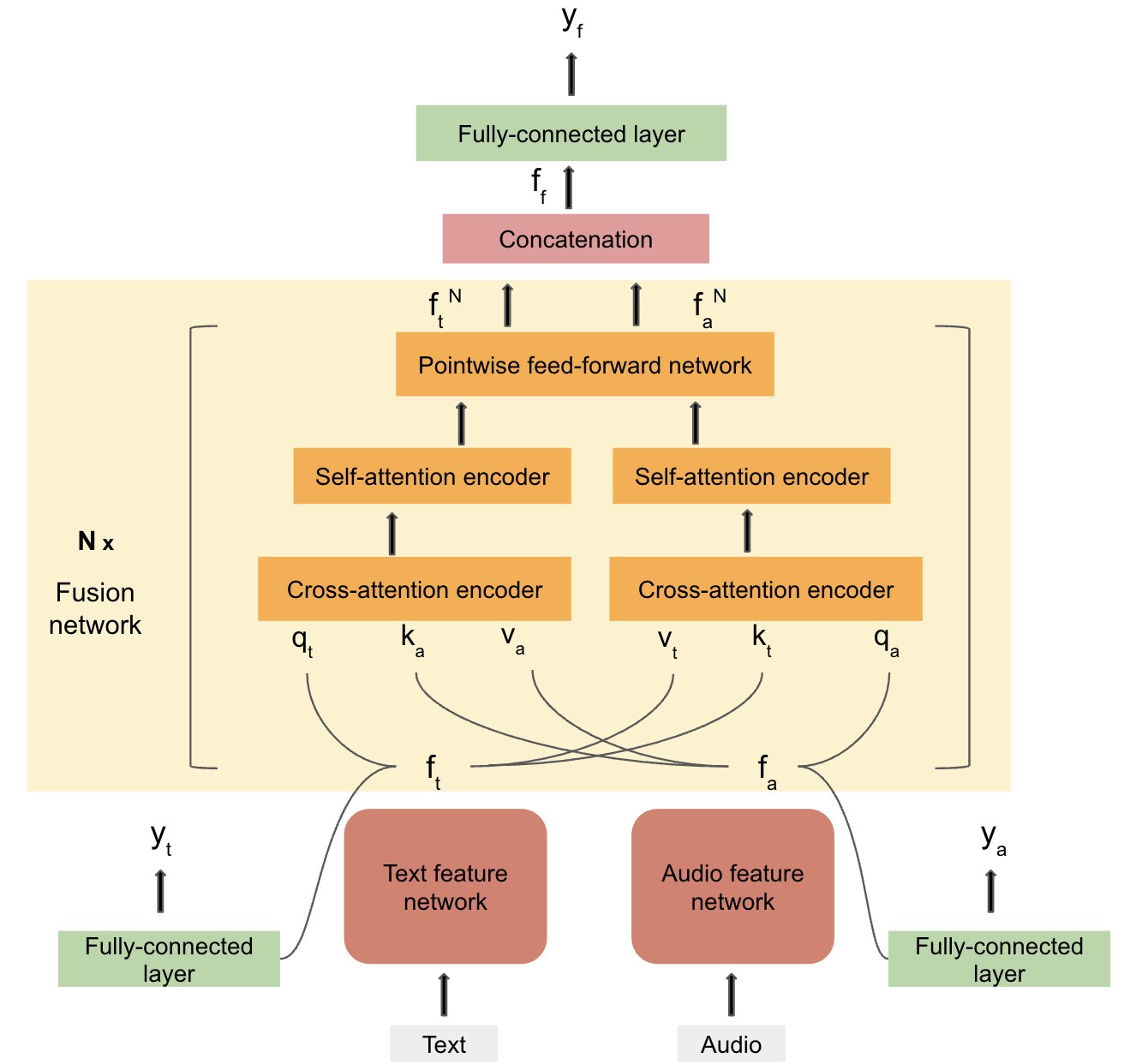}
    \caption{Our Model Structure}
    \label{fig:label1}
  \end{minipage}\hfill
  \begin{minipage}{0.5\textwidth}
    \centering
    \includegraphics[width=1\textwidth]{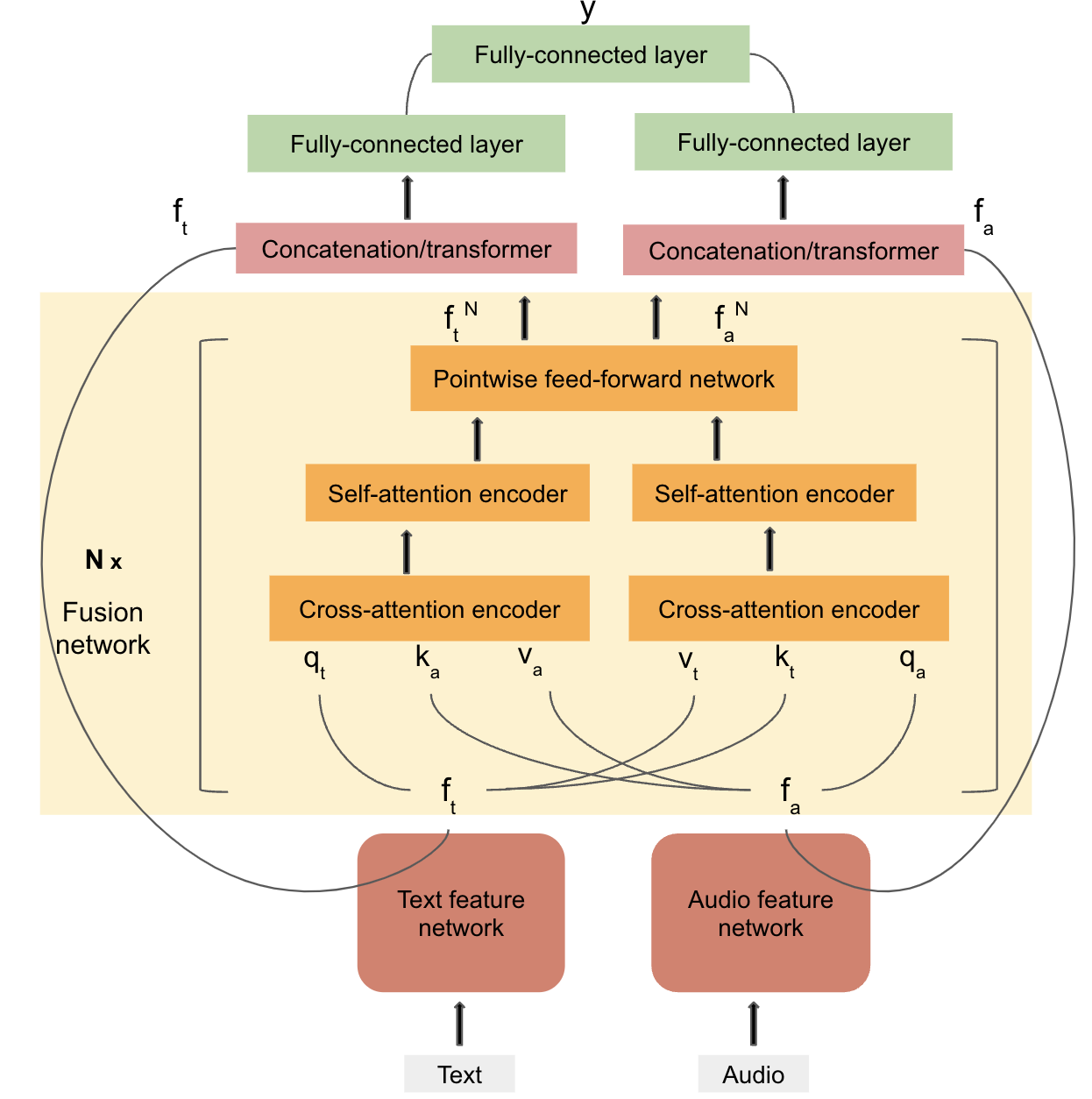}
    \caption{Model Variations}
    \label{fig:label2}
  \end{minipage}
\end{figure*}

\section{Methodology}
% Sara: add more citation, to better differentiate what are the prior work versus our work/methods 
% Sara: seperate methods vs. experiments details.
% Jeff: also mention feature selection and multi-loss training here
The model for sentiment detection in our study involves two primary components: the feature network and the fusion network. Each of these has its own unique mechanisms and contributes towards the overall functioning of our proposed Multi-Modality Multi-Loss Fusion Network (MMML) as illustrated in Figure 1. Additionally, We adopted multi-loss training, experimented with restoring original signals, and explored context modeling.

\subsection{Feature Network}
The Feature Network employs two different pre-trained models for text and audio processing. The text subnet leverages RoBERTa \cite{liu2019roberta}, chosen for its significantly superior performance on various downstream tasks. The audio subnet employs different models for different languages: HuBERT \cite{hsu2021hubert} for Mandarin and Data2Vec \cite{baevski2022data2vec} for English. This ensures the optimized extraction of features from the given modalities, setting a solid foundation for the subsequent fusion process. 

The details and results of the feature network selection process are in Appendix A.5, and the experiments that lead to the exclusion of the vision modality are in Appendix A.3 and A.4.
\subsection{Fusion Network}
% Jeff: shorten the methodology descriptions and put the detailed implementation in the appendix
% Jeff: adjust the formula to avoid confusion and redundancy 
% Jeff: Enhancing the explanation can be achieved significantly by establishing direct links between the equations presented in Section 3.2 and Figure 1. This approach would contribute to a more coherent and visually supported understanding of the network flow.
The Fusion Network is the heart of the MMML, where the information from multiple modalities is combined. This network is divided into three smaller components as shown in the yellow portion of Figure~\ref{fig:label1}. 

First, there is a Cross-Attention Encoder, which adopts a mechanism similar to the self-attention encoder but which employs a query from one modality and uses keys and values generated from another modality. This cross-modal interaction aims to capture the inter-dependencies between different modalities, contributing to a more holistic understanding of the data. This encoder is defined as:

\newenvironment{shrinkeq}[2]%save space for formula
{ \bgroup
  \addtolength\abovedisplayshortskip{#1}
  \addtolength\abovedisplayskip{#1}
  \addtolength\belowdisplayshortskip{#2}
  \addtolength\belowdisplayskip{#2}}
{\egroup\ignorespacesafterend}

\begin{shrinkeq}{-1ex}{-3ex}
{\small
    \begin{equation*}
    \text{Attention}(Q_{m1}, K_{m2}, V_{m2}) = \text{softmax}\left(\frac{Q_{m1} K_{m2}^T}{\sqrt{d_k}}\right)V_{m2}
    \end{equation*}
}
\end{shrinkeq}\\

where:   

\begin{shrinkeq}{-1ex}{-3ex}
{\small
    \begin{align*}
    Q_{m1} &= W_q \cdot f_{m1}\\
    K_{m2} &= W_k \cdot f_{m2}\\
    V_{m2} &= W_v \cdot f_{m2}
    \end{align*}
}
\end{shrinkeq}\\

% \begin{shrinkeq}{-1ex}{-3ex}
% {\small
% \begin{equation*}
% \text{Attention}(Q, K, V) = \text{softmax}\left(\frac{Q K^T}{\sqrt{d_k}}\right)V
% \end{equation*}
% }
% \end{shrinkeq}\\
% where Q is the matrix of queries, K is the matrix of keys, V is the matrix of values, and $d_k$ is the dimension of the keys.\noindent\ignorespacesafterend

We denote queries as $Q_{m1}$ (from modality 1) and the keys and values as $K_{m2}$ and $V_{m2}$ (from modality 2). $f_{m1}$ is the feature from modality 1 and $f_{m2}$ is the feature from modality 2. In Figure 1, the text feature $f_t$ is utilized to build the query $q_t$ for text and is also used to construct the key $k_a$ and value $v_a$ for audio. The cross-attention encoder essentially projects the hidden states from one modality into the space of another modality.

Also, our proposed network includes additional {\it Self-Attention Encoders}. The self-attention mechanism was originally designed to find the correlation within a single modality, thereby capturing the intra-modal dynamics of the data. In our model, the self-attention module serves to model the connections across time steps of the new feature representation after passing through the cross-modality encoder.

Finally, our model includes a {\it Pointwise Feed-Forward Network} which applies fully connected feed-forward networks and ReLU activation functions to each individual position, further refining the encoded feature representations. Through combining these methodologies, we aim to optimize the multi-modal feature extraction and fusion process, enhancing the MMML's performance in sentiment detection tasks.

\subsection{Multi-Loss Training}
% Jeff: The explanation of multi-loss training requires improvement, particularly in elucidating how y_t and y_a are integrated into the final loss function. It remains unclear whether classification is conducted with these vectors, and if so, how the corresponding errors are measured. Additionally, the nature of the final loss, and whether it is the summation of errors between the three vectors, needs clarification. Including an equation depicting the final loss function could significantly enhance the overall explanation.
In order to leverage {\it multi-loss training}, we modified the architecture of our fusion network to incorporate an additional fully-connected layer at the termination of each feature network, as illustrated in the green portion of Figure 1. This modification enables two additional outputs from individual modalities, in addition to the combined feature output. This design facilitates the application of three distinct loss functions during training, each corresponding to one of the outputs. Despite the classification being conducted using only the output from the fusion network, the final loss is the summation of losses from the subnets and the loss from the fusion network:

\begin{shrinkeq}{-1ex}{-3ex}
{\small
    \begin{align*}
    Loss = \sum_{m\in\{a,t,f\}} {\alpha_m * loss\_fn(y_m, target_m)}
    \end{align*}
}
\end{shrinkeq}\\

The source of targets from different modalities is explained in section 4.5 and we witness no significant boost in performance by adjusting $\alpha$ to be other than 1.

The rationale behind implementing additional losses for each individual modality is to bolster the respective feature networks' comprehension and processing of their respective signals. Given that each feature network perceives and handles signals distinctively, akin to how humans discern emotions through different sensory signals, the multi-task loss serves a dual purpose: first, it encourages each feature network to refine its method of processing its specific modality, akin to honing the "sense" associated with that modality; second, it trains the fusion network to effectively combine the distinct signals relayed by the feature networks, as guided by the loss from the combined modality.

Through this multi-loss training approach, we create a model that efficiently mirrors human-like multi-modal emotion perception, each modality working independently and collaboratively to understand the comprehensive emotional context.
% Sara: should we mention mimicing human understanding in the model/loss design in intro/conclusion as well, when talking about our model? As an addtion to how we achieved sota?
% Sara: point out where in the Figure 1

\subsection{Original Signal Restoration}
% Jeff: reduce this part
In our investigation of the fusion network, we developed two variants designed to mitigate potential loss of original signals during the cross-modal projection process, as shown in Figure 2. Because the cross-attention mechanism projects one modality into another, some original signals might be obscured or lost. Therefore, these variations aim to combine the original signal with the projected signal, thereby enhancing the ability of the network to learn from both signals simultaneously.

The first is {\it Concatenation Variation}, which concatenates the original feature with the fused feature. The second is the {\it Transformer Variation}. This variation merges the original hidden states and the fused hidden states along the feature dimension and uses transformer encoders to further process these combined hidden states.

This fusion of original and projected information within each modality aims to maintain the integrity of the original signals, while also integrating the enriched cross-modal information. The combined features then go through a linear layer and are subsequently concatenated with features from other modalities.  

\subsection{Context Modeling}
% We explored the integration of contextual data into existing model frameworks, specifically contrasting two distinct methodologies for context integration. These methodologies were: (i) the \textbf{concatenation} of context and the current utterance as a singular input stream to the model, and (ii) the \textbf{independent processing} of context and current utterance, followed by a subsequent fusion of their respective representational outputs. 

We explored the integration of contextual data (previous utterances) into existing model frameworks, specifically contrasting two distinct methodologies for context integration. These methodologies were: (i) the \textbf{concatenation} of context (previous utterances) and the current utterance as a singular input stream to the model, and (ii) \textbf{independent processing} of context (previous utterances) and current utterance, followed by a subsequent fusion of their respective representational outputs. The first method treats the concatenated input as a single utterance and gives one representation, while the second processes the context and current input separately and gives one representation for each.

\section{Experiments}
\begin{table*}[ht]
% \begin{table*}
\centering
\begin{subtable}{\textwidth}
\centering
\small
\setlength{\tabcolsep}{3pt}
\renewcommand{\arraystretch}{1.2}
\resizebox{\textwidth}{!}{
\begin{tabular}{llllllllllllllll}
\multicolumn{1}{c}{Model} & \multicolumn{7}{c}{\textbf{CMU-MOSI}}                                                                                                             & \multicolumn{1}{l|}{} & \multicolumn{7}{c}{\textbf{CMU-MOSEI}}                                                                                                            \\ \cline{2-8} \cline{10-16} 
                          & $\mathbf{ACC_{2Has0}}$ & $\mathbf{F1_{Has0}}$ & $\mathbf{ACC_{2Non0}}$ & $\mathbf{F1_{Non0}}$ & $\mathbf{ACC_7}$ & $\mathbf{MAE}$ & $\mathbf{Corr}$ & \multicolumn{1}{l|}{} & $\mathbf{ACC_{2Has0}}$ & $\mathbf{F1_{Has0}}$ & $\mathbf{ACC_{2Non0}}$ & $\mathbf{F1_{Non0}}$ & $\mathbf{ACC_7}$ & $\mathbf{MAE}$ & $\mathbf{Corr}$ \\ \cline{1-8} \cline{10-16} 
LMF                       & -                      & -                   & 82.5                   & 82.4                & 33.20            & 0.917          & 0.695           &                       & -                      & -                   & 82.0                   & 82.1                & 48.00            & 0.623          & 0.700           \\
TFN                       & -                      & -                   & 80.8                   & 80.7                & 34.90            & 0.901          & 0.698           &                       & -                      & -                   & 82.5                   & 82.1                & 50.20            & 0.593          & 0.677           \\
MFM                       & -                      & -                   & 81.7                   & 81.6                & 35.40            & 0.877          & 0.706           &                       & -                      & -                   & 84.4                   & 84.3                & 51.30            & 0.568          & 0.703           \\
MTAG                      & -                      & -                   & 82.3                   & 82.1                & 38.90            & 0.866          & 0.722           &                       & -                      & -                   & -                      & -                   & -                & -              & -               \\
SPC                       & -                      & -                   & 82.8                   & 82.9                & -                & -              & -               &                       & -                      & -                   & 82.6                   & 82.8                & -                & -              & -               \\
ICCN                      & -                      & -                   & 83.0                   & 83.0                & 39.00            & 0.862          & 0.714           &                       & -                      & -                   & 84.2                   & 84.2                & 51.60            & 0.565          & 0.704           \\
MuIT                      & 81.50                  & 80.60               & 84.10                  & 83.90               & -                & 0.861          & 0.711           &                       & -                      & -                   & 82.5                   & 82.3                & -                & 0.580          & 0.713           \\
MISA                      & 80.79                  & 80.77               & 82.10                  & 82.03               & -                & 0.804          & 0.764           &                       & 82.59                  & 82.67               & 84.23                  & 83.97               & -                & 0.568          & 0.717           \\
COGMEN                    & -                      & -                   &                        & 84.34               & 43.90            & -              & -               &                       & -                      & -                   & -                      & -                   & -                & -              & -               \\
Self-MM                   & 84.00                  & 84.42               & 85.98                  & 85.95               & -                & 0.713          & 0.798           &                       & 82.81                  & 82.53               & 85.17                  & 85.30               & -                & 0.530          & 0.765           \\
MAGBERT                  & 84.20                  & 84.10               & 86.10                  & 86.00               & -                & 0.712          & 0.796           &                       & 84.70                  & 84.50               & -                      & -                   & -                & -              & -               \\
MIMM                      & 84.14                  & 84.00               & 86.06                  & 85.98               & 46.65            & 0.700          & 0.800           &                       & 82.24                  & 82.66 4             & 85.97                  & 85.94               & 54.24            & 0.526          & 0.772           \\
TEASEL                    & 84.79                  & 84.72               & 87.5                   & 85                  & 47.52            & 64.4           & 83.6            &                       & -                      & -                   & -                      & -                   & -                & -              & -               \\
SPECTRA                   & -                      & -                   & 87.5                   & -                   & -                & -              & -               &                       & -                      & -                   & 87.34                  & -                   & -                & -              & -               \\
UniMSE                    & 85.85                  & 85.83               & 86.9                   & 86.42               & 48.68            & 69.1           & 80.9            &                       & 85.86                  & 85.79               & 87.5          & 87.46     & 54.39            & 52.3           & 77.3            \\ \cline{1-8} \cline{10-16} 
\textbf{MMML}      & 85.91                  & 85.85               & 88.16                  & 88.15               & 48.25            & 64.29          & 83.8            &                       & 86.32         & 86.23     & 86.73                  & 86.49               & 54.95   & 51.74 & 79.08 \\
\textbf{+ context}        & \textbf{87.51}         & \textbf{87.45}      & \textbf{89.69}         & \textbf{89.67}      & \textbf{50.34}   & \textbf{58.31} & \textbf{86.93}  &                       & \textbf{87.24}              & \textbf{87.18}           &             \textbf{88.02}           &       \textbf{88.15}              &        \textbf{55.74}          &      \textbf{49.22}          &      \textbf{81.37}           \\ \cline{1-8} \cline{10-16} 
\end{tabular}
}
\caption{CMU-MOSI and CMU-MOSEI}
\label{tab:CMU-MOSEI1}
\end{subtable}

\vspace{0.3cm}

\begin{subtable}{\textwidth}
\centering
\scriptsize
\setlength{\tabcolsep}{3pt}
\renewcommand{\arraystretch}{1.2}
\begin{tabular}{lllllll}
\hline
 & $\mathbf{ACC_2}$ & $\mathbf{ACC_3}$ & $\mathbf{ACC_5}$ & $\mathbf{F1}$ & $\mathbf{MAE}$ & $\mathbf{Corr}$ \\ \hline
EMT & 80.1 & 67.4 & 43.5 & 80.1 & 39.6 & 62.3 \\ \hline
\textbf{MMML(ours)} & \textbf{82.93} & \textbf{69.37} & \textbf{49.38} & \textbf{82.9} & \textbf{33.2} & \textbf{73.26} \\ \hline
\end{tabular}
\caption{CH-SIMS}
\label{tab:CH-SIMS1}
\end{subtable}
\caption{\textbf{Comparison with SOTA}: Achieved best performance on three datasets. All experimental results presented are averages derived from three separate runs. The performances of baselines are shared by their authors.}
\label{tab:Performance1}
\end{table*}

\subsection{Experimental Setup}
\subsubsection{Datasets}
We use three primary datasets, each characterized by its unique properties and content, to test the performance of the Multi-Modality Multi-Loss Fusion Network on sentiment detection.  

The {\it CMU-Multimodal Opinion Sentiment and Emotion Intensity (CMU-MOSI)} \cite{zadeh2016mosi}: This dataset, developed in English, includes audio, text, and video modalities compiled from 2199 annotated video segments collected from YouTube monologue movie reviews. It offers a focused approach to studying sentiment detection within the context of film critiques.   

The {\it CMU-Multimodal Sentiment Analysis (CMU-MOSEI)} \cite{bagher-zadeh-etal-2018-multimodal} is an extension of CMU-MOSI, incorporating the same modalities of audio, text, and video from YouTube videos, but it has a broader scope, covering a wider range of topics, and is more substantial in size, with 23,453 annotated video segments.

The {\it Chinese Multimodal Sentiment Analysis Dataset (CH-SIMS)} \cite{yu-etal-2020-ch} includes the same modalities in Mandarin: audio, text, and video, collected from 2281 annotated video segments. It includes data from TV shows and movies, making it culturally distinct and diverse, and provides multiple labels for the same utterance based on different modalities, which adds an extra layer of complexity and richness to the data.

These datasets provide a broad and multi-cultural perspective on sentiment detection, allowing for a thorough evaluation and comparative analysis of the MMML's performance across diverse data landscapes.

\subsubsection{Baseline Models}
% [adding paragraphs on Baselines]
% TODO: add citations and below 

In our comprehensive evaluation of the MMML model, we conducted a detailed comparison with a wide array of baseline models in multimodal sentiment analysis. This comparison included several categories of models, each representing a unique approach to multimodal learning.

The first category consisted of early multimodal fusion methods, including Tensor Fusion Network (TFN) \cite{zadeh2017tensor}, Low-rank Multimodal Fusion (LMF) \cite{liu-etal-2018-efficient-low}, and Multimodal Factorization Model (MFM) \cite{DBLP:journals/corr/abs-1806-06176}. These models are foundational in early multimodal fusion approaches to multimodal analysis.

The second category focused on methods that enhance multimodal integration through more modern modality interaction modeling methods. This includes the multimodal Transformer (MulT) by \citet{tsai-etal-2019-multimodal}, Interaction Canonical Correlation Network (ICCN) by \citet{DBLP:journals/corr/abs-1911-05544}, Sparse Phased Transformer (SPC) by \citet{DBLP:journals/corr/abs-2109-12932}, and Modal-Temporal Attention Graph (MTAG) by \citet{yang-etal-2021-mtag}. These models represent a significant advancement in handling complex multimodal interactions.

The third category encompasses models that prioritize modality consistency and difference. This includes MISA \cite{hazarika2020misa}, which manages modal representation space, Self-MM \cite{yu2021le} which leverages multi-task learning from unimodal representations, MAG-BERT \cite{rahman-etal-2020-integrating} with its innovative fusion gate, and MMIM \cite{han-etal-2021-improving} which focuses on maximizing mutual information hierarchically. We also evaluated models focusing on self-supervised learning Transformers for combined modalities, such as TEASEL \cite{arjmand2021teasel}, and those exploring speech-text alignment like SPECTRA \cite{yu2023speechtext}.

Further broadening our comparison, we included models like COGMEN \cite{joshi-etal-2022-cogmen}, which consider the multimodal conversational context. A noteworthy inclusion was UniMSE \cite{hu-etal-2022-unimse}, a model proposing a knowledge-sharing framework that unifies multimodal sentiment analysis and emotion recognition in conversation tasks, currently regarded as the state-of-the-art (SOTA) method.

In comparing the feature extractor between ours and the baselines, we find our approach to be on par with the baselines in terms of selecting feature extractors. Like ours, the majority of the baseline models employ advanced pre-trained models for feature extraction. For example, TEASEL adds speech tokens on top of a pre-trained RoBerta during training, which also uses advanced pre-trained models as the feature extractor, which is comparable to ours.

Our methods further explore the effectiveness of cross-modality attention, aiming to further optimize the multimodal representation during the projection of one modality into another, along with enhancing to select optimal training methods including multi-loss training and context-modeling, thus enriching our ability to process and understand multimodal data.

\subsubsection{Metrics}
Our MMML model was evaluated using metrics consistent with existing research against existing benchmarks, which enables comprehensive evaluation of our model's performance across diverse sentiment analysis dimensions (detailed descriptions in Appendix A.6). Additional sets of ablation experiments on different components of the model were conducted for analysis, to interpret and explain the model performance. 
% Sara: add experimental set-up and overall explination of the experiments ran?

\subsection{Results}
Our overall results are shown in Table 1. When compared with contemporary state-of-the-art models, our method emerges as a robust performer, offering superior outcomes for both CMU-MOSI and CMU-MOSEI. Among recent models, UniMSE \cite{hu-etal-2022-unimse} has delivered the best results on the English datasets. Nonetheless, our MMML model surpasses UniMSE in most of the evaluation metrics, reinforcing the effectiveness of our approach. Adding context further elevates performance substantially for all metrics.

%Intriguingly, there is little research on CH-SIMS for sentiment detection tasks. However
For CH-SIMS, one state-of-the-art model is the Efficient Multimodal Transformer (EMT) \cite{Sun_2023}, which has demonstrated a high degree of performance over existing methods. Our MMML model also significantly outperforms EMT across all metrics, further underscoring the potential of our fusion network.

We also note that our model uses only audio and text signals, while these other models take advantage of all three signals. This further proves the effectiveness of our model.
These results not only validate our multimodal fusion network but also affirm the robustness of our chosen methodology for sentiment detection tasks. Our impressive performance on all three datasets, CMU-MOSI, CMU-MOSEI, and CH-SIMS, verifies the versatility and adaptability of our MMML model, emphasizing its value in advancing the field of sentiment detection.
\begin{table}
\small
\begin{subtable}{0.5\textwidth}
\centering
\begin{tabular}{ll}
\hline
\textbf{Feature name} & $\mathbf{ACC_2}$ \\ \hline
openSMILE             & 0.6696               \\ 
Mel Spectrogram       & 0.6805               \\ 
Fine-tuned HuBert(CH) & \textbf{0.7465}       \\ \hline
\end{tabular}
\caption{CH-SIMS}
\label{tab:dataset2}
\end{subtable}

\vspace{0.3cm}

\begin{subtable}{0.5\textwidth}
\centering
\begin{tabular}{ll}
\hline
\textbf{Feature name} & $\mathbf{ACC_2}$ \\ \hline
openSMILE             & 0.4606               \\ 
Mel Spectrogram       & 0.4519               \\ 
Fine-tuned Data2vec(EN) & \textbf{0.7099}     \\ \hline
\end{tabular}
\caption{CMU-MOSI}
\label{tab:dataset1}
\end{subtable}
\caption{\textbf{Audio Feature Selection Results}: Fine-tuning a pre-trained audio model works significantly better than using other audio features.}
\label{tab:Performance2}
\end{table}

\subsection{Audio Feature Comparison}
% Jeff: add other audio pre-trained models
% Jeff: add a section for text feature selection?
To incorporate the best speech information into our model, the initial stage of our experimentation process involved comparing the performance on audio features for sentiment analysis from two datasets, CMU-MOSI (English) and CH-SIMS (Mandarin), using openSMILE and Mel spectrograms, each with customized parameters for optimal feature extraction to compare with features from a pre-trained audio model (details in Appendix A.2).

Upon evaluation of the different audio feature extraction methods shown in Table 2, we found that use of a pre-trained model for raw audio yielded higher accuracy rates: accuracy rates of approximately 71\% and 75\% were achieved for CMU-MOSI and CH-SIMS, respectively. This outperformed the other two hand-crafted features (openSMILE and Mel spectrograms) by a significant margin. Interestingly, openSMILE and Mel spectrograms displayed comparable performance on CH-SIMS. However, their performance on CMU-MOSI was notably subpar. We hypothesize that CH-SIMS, comprising audio from TV shows and movies, presents a more straightforward task for audio sentiment detection.

This analysis highlights the effectiveness of using pre-trained models for raw audio in achieving superior sentiment classification accuracy. It also underscores the need to consider the characteristics and source of audio data in applying different feature extraction techniques.

\subsection{Fusion Network Ablation Experiment}
In our ablation study focusing on the fusion network with the CMU-MOSI dataset, we identified crucial components that significantly contribute to the network's performance. Specifically, the self-attention layers and the three fully connected layers following the cross-attention layers proved to be vital, as demonstrated in Table 8 of Appendix A.7.

Upon removing the self-attention layers, we observed a noticeable decline in model performance across all metrics. This decline became more pronounced when both the self-attention and fully connected layers were eliminated. These findings underscore the importance of these components in enhancing and refining modality representations within the fusion network, highlighting their integral role in our model's architecture.

\subsection{Comparison of Simple Concatenation and Fusion Network}
% Jeff: the efficacy of the MMML fusion network approach in comparison to solely relying on cross-modal attention remains unclear. To elucidate the advantages of the MMML fusion network, it would be beneficial to incorporate a new experiment exclusively employing cross-modal attention—omitting the use of self-attention and Pointwise Feed-Forward Network. This additional experiment would provide valuable insights into the specific benefits derived from the MMML fusion network.
To demonstrate the superiority of our proposed fusion network, we compared it to concatenation. Upon analysis of our results (Table 3), we observed that the introduction of the transformer fusion network yielded improvements in performance in most metrics for CMU-MOSEI and CH-SIMS, and for half of the metrics for CMU-MOSI. These results underscore the effectiveness of our transformer fusion network in enhancing cross-modality modeling and suggest its potential as a powerful tool for multi-modal sentiment detection.

Beyond these observations, it is important to highlight that both methods which combined audio and text signals outperformed methods utilizing only text signals in all metrics across the three datasets. A noteworthy increase in performance was recorded on the CH-SIMS dataset upon the addition of audio signals, while the two English datasets, CMU-MOSI and CMU-MOSEI, exhibited smaller improvements. The substantial improvement observed in CH-SIMS can be attributed to two factors. First, CH-SIMS assigns unique labels to audio and text, thereby facilitating the network's ability to learn distinct signals from each modality. Second, the source for CH-SIMS is TV show and movie videos, which typically display easily-interpretable sentiments. This characteristic probably contributes to the effectiveness of combining audio and text signals for sentiment detection.

\subsection{Multi-loss Training Experiments}
To investigate the effectiveness of multi-loss training, we performed comparative experiments on two different datasets: CMU-MOSEI and CH-SIMS. CMU-MOSEI provides a single target for each utterance, whereas CH-SIMS offers different labels for each modality in addition to the combined modalities.
We easily adapted multi-loss training to CH-SIMS, given its distinct labels for each modality. For CMU-MOSEI, we duplicated the single target across different losses to enable multi-task training. 

The results, as shown in Table 4, were striking: while multi-loss and single-loss training performed similarly on CMU-MOSEI, multi-loss training significantly boosted performance on CH-SIMS. This underscores the value of unique labels for each modality when employing multi-loss training. The improved performance on CH-SIMS can be attributed to the distinct nature of the signals processed by each feature network. Since audio includes acoustic signals that are not present in the text, it is common for them to produce different sentiments. Having distinct labels assists each network in learning better how to process its unique signal.

Surprisingly, as shown in Table 5, the multi-loss training also contributed to an enhanced performance of the text subnet when compared to training with only the text. The additional audio signal appears to support the performance improvement. This suggests that even when the goal is to use only the text input for inference, multi-loss training can be beneficial. The text subnet, after training with the multi-modal model, can be extracted and used independently, offering superior performance compared to being trained alone.

Interestingly, this improvement was not observed in the audio subnet, potentially due to the stronger signal from the text subnet (reflected by a 10\% higher accuracy when trained alone) which made it easier to train, and thus the network might have focused on reducing its loss.

In summary, the benefits of multi-loss training are threefold. First, it substantially boosts the performance of the entire network when distinct labels for different modalities are available. Second, loss from other modalities enhances the performance of the text subnet, indicating that we can utilize other modalities in training even when the text subnet is the only required component for inference. Third, it is capable of handling missing modalities, enabling outputs when only text or audio inputs are available. These findings shed light on the potential of multi-loss training in the context of multi-modality fusion networks, opening avenues for further research and optimization.
\begin{table*}
\centering
\begin{subtable}{\textwidth}
\centering
\small
\setlength{\tabcolsep}{3pt}
\renewcommand{\arraystretch}{1.2}
\begin{tabular}{lllllllll}
\hline
 & $\mathbf{Has0\_ACC_2}$ & $\mathbf{Has0\_F1}$ & $\mathbf{Non0\_ACC_2}$ & $\mathbf{Non0\_F1}$ & $\mathbf{ACC_5}$ & $\mathbf{ACC_7}$ & $\mathbf{MAE}$ & $\mathbf{Corr}$ \\ \hline
\textbf{text-only} & 84.89 & 84.86 & 87.04 & 87.07 & 54.81 & 47.32 & 66.62 & 83.27 \\ \hline 
\textbf{concatenation} & 85.77 & 85.74 & 87.6 & 87.62 & \textbf{56.51} & \textbf{48.79} & \textbf{64.27} & \textbf{84.06} \\ 
\textbf{+ fusion network} & \textbf{85.91} & \textbf{85.85} & \textbf{88.16} & \textbf{88.15} & 56.08 & 48.25 & 64.29 & 83.8 \\ \hline
\end{tabular}
\caption{CMU-MOSI}
\label{tab:CMU-MOSI2}
\end{subtable}
\vspace{0.3cm}
\begin{subtable}{\textwidth}
\centering
\small
\setlength{\tabcolsep}{3pt}
\renewcommand{\arraystretch}{1.2}
\begin{tabular}{lllllllll}
\hline
 & $\mathbf{Has0\_ACC_2}$ & $\mathbf{Has0\_F1}$ & $\mathbf{Non0\_ACC_2}$ & $\mathbf{Non0\_F1}$ & $\mathbf{ACC_5}$ & $\mathbf{ACC_7}$ & $\mathbf{MAE}$ & $\mathbf{Corr}$ \\ \hline
\textbf{text-only} & 84.81 & 84.95 & 86.34 & 86.19 & 54.99 & 52.7 & 53.31 & 78.6 \\ \hline
\textbf{concatenation} & 84.77 & 84.9 & \textbf{86.82} & \textbf{86.65} & 55.99 & 53.94 & 51.63 & \textbf{79.81} \\ 
\textbf{+ fusion network} & \textbf{86.32} & \textbf{86.23} & 86.73 & 86.49 & \textbf{57.32} & \textbf{54.95} & \textbf{51.54} & 79.08 \\ \hline
\end{tabular}
\caption{CMU-MOSEI}
\label{tab:CMU-MOSEI2}
\end{subtable}
\vspace{0.3cm}
\begin{subtable}{\textwidth}
\centering
\small
\setlength{\tabcolsep}{3pt}
\renewcommand{\arraystretch}{1.2}
\begin{tabular}{lllllll}
\hline
 & $\mathbf{ACC_2}$ & $\mathbf{ACC_3}$ & $\mathbf{ACC_5}$ & $\mathbf{F1}$ & $\mathbf{MAE}$ & $\mathbf{Corr}$ \\ \hline
\textbf{text-only} & 79.21 & 65.06 & 42.02 & 79.14 & 42.65 & 59.4 \\ \hline
\textbf{concatenation} & 81.91 & \textbf{70.68} & 47.12 & 82.1 & 34.96 & 72.37 \\ 
\textbf{+ fusion network} & \textbf{82.93} & 69.37 & \textbf{49.38} & \textbf{82.9} & \textbf{33.2} & \textbf{73.26} \\ \hline
\end{tabular}
\caption{CH-SIMS}
\label{tab:CH-SIMS2}
\end{subtable}
\caption{\textbf{Concatenation vs.~Transformer Fusion}: Integration of audio signals enhances performance across almost all metrics, with more pronounced impact on CH-SIMS. Implementing the Fusion Network augments performance slightly in most metrics. All experimental results presented are averages derived from three separate runs.}
\label{tab:Performance3}
\end{table*}

\begin{table*}
\centering
\begin{subtable}{\textwidth}
\centering
\small
\setlength{\tabcolsep}{3pt}
\renewcommand{\arraystretch}{1.2}
\begin{tabular}{lllllllll}
\hline
 & $\mathbf{Has0\_ACC_2}$ & $\mathbf{Has0\_F1}$ & $\mathbf{Non0\_ACC_2}$ & $\mathbf{Non0\_F1}$ & $\mathbf{ACC_5}$ & $\mathbf{ACC_7}$ & $\mathbf{MAE}$ & $\mathbf{Corr}$ \\ \hline
\textbf{single-loss} & \textbf{85.22} & \textbf{85.39} & \textbf{87.02} & \textbf{86.91} & 55.95 & 53.85 & 51.96 & 79.68 \\ \hline
\textbf{multi-loss} & 84.77 & 84.9 & 86.82 & 86.65 & \textbf{55.99} & \textbf{53.94} & \textbf{51.63} & \textbf{79.81} \\ \hline
\end{tabular}
\caption{CMU-MOSEI}
\label{tab:CMU-MOSEI3}
\end{subtable}
\vspace{0.3cm}
\begin{subtable}{\textwidth}
\centering
\small
\setlength{\tabcolsep}{3pt}
\renewcommand{\arraystretch}{1.2}
% \resizebox{\textwidth}{!}{%
\begin{tabular}{lllllll}
\hline
 & $\mathbf{ACC_2}$ & $\mathbf{ACC_3}$ & $\mathbf{ACC_5}$ & $\mathbf{F1}$ & $\mathbf{MAE}$ & $\mathbf{Corr}$ \\ \hline
\textbf{Single-loss} & 78.34 & 67.18 & 46.83 & 78.59 & 39.09 & 62.69 \\ \hline
\textbf{multi-loss} & \textbf{81.91} & \textbf{70.68} & \textbf{47.12} & \textbf{82.1} & \textbf{34.96} & \textbf{72.37} \\ \hline
\end{tabular}
% }
\caption{CH-SIMS}
\label{tab:CH-SIMS3}
\end{subtable}
\caption{\textbf{Single-Loss Training vs.~Multi-Loss Training}: While multi-loss training does not yield performance improvement when identical labels are used for different losses, as in the case of CMU-MOSEI, it does contribute significantly to performance enhancement when unique labels are assigned to each modality, as observed with CH-SIMS.}
\label{tab:Performance4}
\end{table*}

\begin{table*}
% \centering
% \begin{subtable}{\textwidth}
% \centering
% \small
% \setlength{\tabcolsep}{3pt}
% \renewcommand{\arraystretch}{1.2}
% \begin{tabular}{lllllllll}
% \hline
%  & $\mathbf{Has0\_ACC_2}$ & $\mathbf{Has0\_F1}$ & $\mathbf{Non0\_ACC_2}$ & $\mathbf{Non0\_F1}$ & $\mathbf{ACC_5}$ & $\mathbf{ACC_7}$ & $\mathbf{MAE}$ & $\mathbf{Corr}$ \\ \hline
% \textbf{text-loss only} & 84.79 & 84.72 & 87.29 & 87.29 & \textbf{56.41} & \textbf{48.68} & 64.96 & 83.61 \\ \hline
% \textbf{multi-loss} & \textbf{85.62} & \textbf{85.56} & \textbf{87.91} & \textbf{87.9} & 55.01 & 47.42 & \textbf{64.76} & \textbf{83.79} \\ \hline
% \end{tabular}
% \caption{CMU-MOSI}
% \label{tab:CMU-MOSI3}
% \end{subtable}
% \vspace{0.3cm}
\begin{subtable}{\textwidth}
\centering
\small
\setlength{\tabcolsep}{3pt}
\renewcommand{\arraystretch}{1.2}
\begin{tabular}{lllllllll}
\hline
 & $\mathbf{Has0\_ACC_2}$ & $\mathbf{Has0\_F1}$ & $\mathbf{Non0\_ACC_2}$ & $\mathbf{Non0\_F1}$ & $\mathbf{ACC_5}$ & $\mathbf{ACC_7}$ & $\mathbf{MAE}$ & $\mathbf{Corr}$ \\ \hline
\textbf{text-loss only} & \textbf{84.81} & \textbf{84.95} & 86.34 & 86.19 & 54.99 & 52.97 & 53.31 & 78.6 \\ \hline
\textbf{multi-loss} & 84.36 & 84.62 & \textbf{86.85} & \textbf{86.76} & \textbf{56.06} & \textbf{53.61} & \textbf{52.35} & \textbf{79.49} \\ \hline
\end{tabular}
\caption{CMU-MOSEI}
\label{tab:CMU-MOSEI4}
\end{subtable}
\vspace{0.3cm}
\begin{subtable}{\textwidth}
\centering
\small
\setlength{\tabcolsep}{3pt}
\renewcommand{\arraystretch}{1.2}
\begin{tabular}{lllllll}
\hline
 & $\mathbf{ACC_2}$ & $\mathbf{ACC_3}$ & $\mathbf{ACC_5}$ & $\mathbf{F1}$ & $\mathbf{MAE}$ & $\mathbf{Corr}$ \\ \hline
\textbf{text-loss only} & 79.21 & 65.06 & 42.02 & 79.14 & 42.65 & 59.4 \\ \hline
\textbf{multi-loss} & \textbf{83.15} & \textbf{72.14} & \textbf{48.21} & \textbf{83.74} & \textbf{28.58} & \textbf{78.72} \\ \hline
\end{tabular}
\caption{CH-SIMS}
\label{tab:CH-SIMS4}
\end{subtable}
\caption{\textbf{Impact of Multi-Loss on Text Subnet}: Utilizing audio-related losses can enhance performance of the text subnet, even when identical labels are employed, as is the case with CMU-MOSEI. Remarkably, using specific labels for different modalities results in a substantial performance boost in the text subnet, as evidenced by the results from CH-SIMS.}
\label{tab:Performance5}
\end{table*}

\subsection{Results for Original Signal Restoration}
% Sara: Let's highlight the observation 1 and 2 in the experiment analysis too, and put them into more details. For example, start with a new paragraph and say how futrue researchers can use observation to design their model? 
% 
To understand the effect of restoring original signals, we conducted a comparative analysis of proposed fusion network variations, which reveals a relatively consistent performance across all variations. As shown by the results presented in Table 6 in Appendix A.7, the three methods demonstrate similar performance across all metrics for both CMU-MOSEI and CH-SIMS. Surprisingly, re-incorporating the original signal into the fused signal did not lead to any significant improvement in performance.
% unfinished: (This unexpected result warrants further investigation. One possible approach to gaining insight into this phenomenon is to examine the weights of the last linear layer. This could help ascertain if the original signal was indeed utilized or if it was largely ignored during the training process.
% Additionally, it could be valuable to explore other aspects of the fusion network architecture. For instance, the attention mechanism, a crucial element of the fusion network, could be analyzed to discern the interplay between original and fused signals. This could be achieved by visualizing attention maps to see if specific patterns or alignments emerge. It may also be insightful to scrutinize the gradients during the backpropagation process to see if certain signals are consistently given more importance.)
In essence, while similar performance across different fusion network variations was unanticipated, it paves the way for a deeper understanding of the interactions within the fusion network and the role of original signals in such approaches.

\subsection{Context Modeling Experiments}
% In our investigation, we sought to explore the integration of contextual information utilizing the CMU-MOSI dataset, characterized by many sequential utterances in dialogues, in contrast to the CH-SIMS dataset. This exploration was initially conducted utilizing solely textual data streams, with variable contextual windows (i.e., the number of preceding utterances considered as contextual input). As shown in Table 7 of Appendix A.7, adding context boosts the model performance significantly. Our empirical findings indicated a superior capacity for context window management in the second methodology that processes context and current utterance separately. Notably, while the optimal performance of the concatenation approach was observed at a window of 1, the separation method exhibited performance enhancements up to a context window of 2. In addition, the separation method demonstrated superior performance across all evaluated metrics.

% We next extended our research to the incorporation of audio signals, examining various permutations of context windows for both textual and auditory data streams. Optimal results were achieved with a textual context window of 2 and an auditory context window of 1. This contextual model markedly outperformed our best non-contextual model across all evaluative metrics as shown in Table 1, signifying a significant advancement in the model's performance capabilities.

In our investigation, we sought to explore the integration of contextual information utilizing the CMU-MOSI dataset, characterized by many sequential utterances in dialogues, in contrast to the CH-SIMS dataset.

This exploration was initially conducted utilizing solely textual data streams to compare the two methods ((i) concatenation and (ii) independent processing), with different contextual window lengths (i.e., the number of preceding utterances considered as contextual input). As shown in Table 7 of Appendix A.7, our empirical findings indicated a superior capacity for context window management in the (ii) methodology that processes context and current utterance separately. Notably, while the optimal performance of the (i) concatenation approach was observed at a window of 1, the (ii) independent processing method exhibited performance enhancements up to a context window of 2. In addition, the (ii) independent processing method demonstrated superior performance across all evaluated metrics.

We next extended our research to the incorporation of audio signals, examining the effect of various permutations of context window lengths of both text and audio inputs. Optimal results were achieved with a textual context window of 2 and an auditory context window of 1. As shown in Table 1, this contextual model markedly outperformed our best non-contextual model across all evaluative metrics, identifying a significant advancement in the model's performance capabilities.

\section{Conclusion}

In conclusion, this study has provided novel, important findings for multi-modal sentiment analysis that should benefit future researchers in the designing of sentiment analysis and other models. First, the use of pre-trained models for raw audio yielded superior results, highlighting their effectiveness in feature extraction. Second, combining audio and text signals consistently outperformed using text signals alone, with the transformer fusion network showing promise in enhancing cross-modality modeling. Third, multi-loss training proved beneficial for performance, particularly with unique labels for each modality. Fourth, context information boosts model performances significantly. Last, achieving state-of-the-art results on three sentiment detection datasets underscores the effectiveness of our approach. 

Moreover, in analysis of ablation studies, we show model performance can be improved through a method that reflects a similar pattern with human understanding of sentiment analysis through the multi-loss training. Moreover, multimodal features improve both the overall model performance in multimodality and in single-modality input settings. We provide more effective ways to handle missing modalities and utilized individual modality representation, by allowing the model to train on multimodal features and boost performance on single modality input using multi-loss training. Still, the performance of fusion network variations did remain consistent, prompting further investigation. 
% Overall, these findings contribute to our understanding of multi-modal emotion detection and open avenues for future research in this area. 
%In the future, we encourage researchers to look into xyz....

% We discovered that training on multimodal features improves single modality testing and designing fusion methods based on dataset annotation schema enhances model performance.

\section*{Acknowledgements}

This research is supported in part by the Defense Advanced Research Projects Agency (DARPA), via the CCU Program contract HR001122C0034. The views, opinions and/or findings expressed are those of the authors and should not be interpreted as representing the official views or policies of the Department of Defense or the U.S. Government.
\clearpage

\section{limitations}
\label{sec:limitations}

One limitation of this paper is that the proposed model is studied on two language, English and Mandarin. Another is that the coverage of domains is limited to the design of the datasets we choose to use, which is from YouTube Videos and TV shows. Hence, it is likely that a portion of the data is acted rather than naturally occurring in real life, and acted emotions may be expressed differently than naturally occurring emotions. Such bias in the dataset might lead to learning similar bias in model features and cause errors in recognition if applied to real life situations, which could be different in characteristics and distribution than YouTube videos and TV shows.

Another limitation is that among the 3 public dataset we used, which are all collected from YouTube and TV shows, not all have detailed descriptions about anonymization of the persons appeared in the dataset. However, we did not modify the dataset, since the datasets are widely used and we would like to create coherent results, comparable with previous work.

As for potential risk of misuse, since the paper is focused on more fundamental aspects of the research, it is possible that the model might not perform well if deployed in other scenarios without additional fine-tuning and training, because the model is trained on public datasets collected from TV shows and YouTube. Misuse of directly deploying the model in real-life applications might create risks that the prediction might not always be accurate. 

Finally, our model does not yet incorporate vision features. In developing the model, we initially experimented with incorporating vision features including openFace features and embeddings from a finetuned VGGFace2 model during the early stages. However, our findings indicated that these features did not significantly enhance performance and required a substantial increase in computational resources --- without a commensurate improvement in results. Importantly, our current model achieves performance metrics that surpass state-of-the-art (SOTA) models which do incorporate vision features. This accomplishment underscores the effectiveness of our approach relying on audio and text. Nonetheless, integrating vision features remains a potential avenue for future development. The exploration of vision capabilities is a promising direction for enhancing our model's performance, particularly in areas where visual context can provide additional insights.

\bibliography{acl}
% \bibliography{anthology, acl}
\bibliographystyle{acl_natbib}

\clearpage

\appendix

\section{Appendix}
\label{sec:appendix}

%JH: do you need this section below?  if so, is it in the right place?
% moved it to appendix
\subsection{Training Details}
The training process employed a learning rate of 1e-5, batch size of 16, and the AdamW optimizer. L2 loss was used to optimize the model during the training process. The validation set loss and accuracy were monitored to ensure that the model was not overfitting to the training data. An early stopping mechanism with patience of 8 epochs was employed to ensure the generalizability of the model. The entire procedure was conducted on a single RTX 4090 GPU.
For the audio pre-trained model, the Convolutional Neural Network (CNN) portion used for feature extraction was frozen. The impact of different learning rates for various parts of the network was explored, but no significant differences were observed. Moreover, we found that using 5 layers of fusion network achieves the best results.
All the results presented in the tables are averaged over three independent runs.
\subsection{Audio Feature Extraction and Modeling Details}
For openSMILE, we manipulated the frame size and step, setting them to 0.06 seconds and 0.02 seconds respectively. For Mel spectrograms, the number of Mel filterbanks was set to 128, while the window size and step were adjusted to 0.06 seconds and 0.02 seconds respectively. These configurations were chosen to enhance the precision of audio feature extraction without sacrificing computational efficiency.

Following the feature extraction phase, these features were used to construct models with varying architectures: Transformer models, incorporating between 2 and 4 encoder layers complemented with positional encoding, were employed to process the openSMILE features. A feed-forward layer was subsequently added to process feature embeddings in the CLS token of the final transformer encoder. For processing Mel spectrogram features, we leveraged convolutional neural network (CNN) models, including a custom 8-layer CNN model and modified versions of ResNet-18 and ResNet-32. The choice of these CNN architectures was driven by their known effectiveness in handling image-like data structures such as spectrograms.

\subsection{Vision Features Experiments}

For facial extraction, we initially employed MTCNN, but switched to the OpenCV DNN model due to its superior performance under dim lighting conditions. We extracted faces at a rate of 5 frames per second.

Our experiments involved two models: 1) a CNN-transformer model, combining a CNN pre-trained on the VGG-face2 dataset for spatial relationships with a transformer for temporal relationships; 2) TimesFormer \cite{bertasius2021spacetime}, a transformer-based model pre-trained on video data. Both models achieved about 70\% accuracy on the CH-SIMS dataset (0.7045 for CNN-transformer and 0.7294 for TimesFormer). However, their performance on CMU-MOSI was significantly lower (40-50\% accuracy), leading us to conclude that the facial expressions in CMU-MOSI lack the distinct emotional information necessary for effective sentiment analysis.

\subsection{Modality Selection}

Our model's foundation is a text-based Large Language Model (LLM), while the other two modalities serve as auxiliary signals. Standalone, the text LLM, exhibits over 84 percent accuracy in English datasets and nearly 80 percent accuracy in the Mandarin dataset, outperforming other modalities in all evaluated metrics. We explored combinations of text with audio or video. The text-audio combination yielded a significant performance boost (see Table 3), whereas the text-video combination was less consistently beneficial. Specifically, video signals provide slight performance improvement in the CH-SIMS dataset, but not in the CMU-MOSI dataset. Our analysis revealed that CMU-MOSI videos contain more restrained and ambiguous facial expressions, making them less useful for sentiment detection compared to the emotionally richer expressions in CH-SIMS, sourced from TV shows and movies.

Given these findings, we chose not to include vision features in our final model. This decision was driven by two factors: 1) the inconsistent performance benefits, dependent on the dataset; 2) the substantial computational resources required for processing visual data (details explained in Appendix A.3). This latter aspect not only increases computational overhead but also introduces delays in real-time inference, without a corresponding improvement in results. We recognize that our approach lacks scalability in integrating multiple modalities due to the necessity of large pre-trained models for each to obtain the best performance.

Importantly, our model still surpasses state-of-the-art models that include vision features in performance metrics, emphasizing the efficacy of our text and audio-based approach. Nevertheless, the potential for integrating vision features remains a promising area for future enhancements. The utility of visual context, especially in scenarios where it provides critical insights, warrants further exploration.

\subsection{Feature Network Selection}
In our sentiment analysis framework, we conducted extensive experiments with various text and audio models. For text processing, we evaluated MacBert, Bert, and RoBerta across both Mandarin and English datasets. RoBerta consistently outperformed the others in both languages, demonstrating its superior efficacy in understanding and processing textual data.

For the audio component in the English dataset, we tested three Automatic Speech Recognition (ASR) models: wave2vec, HuBert, and Data2Vec. Among these, Data2Vec emerged as the most effective, providing the best performance in terms of accuracy and reliability. As there was no fine-tuned version of Data2Vec for Mandarin that is publicly available, we compared the performance of wave2vec and HuBert, ultimately selecting HuBert for its superior performance with Mandarin. An interesting observation from our experiments was the positive correlation between the performance of ASR models in sentiment detection and their word error rate (WER) in public benchmarks.
%JH: so no WhisperX?

\subsection{Metrics}
Our MMML model was evaluated using metrics consistent with existing research. 

For CMU-MOSI and CMU-MOSEI, we used: 
\begin{itemize}
\item $\mathbf{Has0\_ACC_2}$, $\mathbf{Has0\_F1}$, including zero sentiment scores as positive; 
\item $\mathbf{Non0\_ACC_2}$, $\mathbf{Non0\_F1}$, ignoring zero sentiment scores;
\item $\mathbf{ACC_5}$, $\mathbf{ACC_7}$, representing 5-class and 7-class accuracies respectively; 
\item $\mathbf{MAE}$, Mean Absolute Error; 
\item $\mathbf{Corr}$, assessing the correlation between predicted and actual scores.
\end{itemize}

For CH-SIMS, we utilized:
\begin{itemize}
\item $\mathbf{ACC_2}$, $\mathbf{ACC_3}$, $\mathbf{ACC_5}$, representing 2-class, 3-class, and 5-class accuracies respectively; 
\item $\mathbf{F1}$, balancing precision and recall; 
\item $\mathbf{MAE}$, mean absolute error; 
\item $\mathbf{Corr}$, assessing correlation between predicted and actual scores.
\end{itemize}
These metrics enable comprehensive evaluation of our model's performance across diverse sentiment analysis dimensions.

\subsection{Additional tables}

\begin{table*}
\centering
\begin{comment}
\begin{subtable}{\textwidth}
\centering
\begin{tabular}{|l|l|l|l|l|l|l|l|l|}
\hline
 & \textbf{Has0\_acc\_2} & \textbf{Has0\_F1} & \textbf{Non0\_acc\_2} & \textbf{Non0\_F1} & \textbf{acc\_5} & \textbf{acc\_7} & \textbf{MAE} & \textbf{Corr} \\ \hline
\textbf{fused features only} & \textbf{85.91} & \textbf{85.85} & \textbf{88.16} & \textbf{88.15} & 56.08 & 48.25 & 64.29 & \textbf{83.8} \\ \hline
\textbf{concat} & 85.28 & 85.23 & 87.75 & 87.76 & \textbf{57.19} & \textbf{49.18} & 65.81 & 82.88 \\ \hline
\textbf{transformer} & 85.18 & 85.13 & 87.35 & 87.35 & 56.36 & 48.93 & \textbf{64.03} & 83.03 \\ \hline
\end{tabular}
\caption{CMU-MOSI}
\label{tab:CMU-MOSI4}
\end{subtable}
\end{comment}
\begin{subtable}{\textwidth}
\centering
\small
\setlength{\tabcolsep}{3pt}
\renewcommand{\arraystretch}{1.2}
\begin{tabular}{lllllllll}
\hline
 & $\mathbf{Has0\_ACC_2}$ & $\mathbf{Has0\_F1}$ & $\mathbf{Non0\_ACC_2}$ & $\mathbf{Non0\_F1}$ & $\mathbf{ACC_5}$ & $\mathbf{ACC_7}$ & $\mathbf{MAE}$ & $\mathbf{Corr}$ \\ \hline
\textbf{Fused Features Only} & \textbf{86.32} & \textbf{86.23} & 86.73 & 86.49 & \textbf{57.32} & 54.95 & \textbf{51.54} & 79.08 \\ \hline
\textbf{Concatenation} & 84.96 & 85.09 & \textbf{86.78} & \textbf{86.61} & 56.86 & \textbf{57.78} & 51.88 & \textbf{79.09} \\ \hline
\textbf{Transformer} & 86.11 & 86.08 & 86.7 & 86.46 & 57.01 & 54.31 & 51.97 & 78.96 \\ \hline
\end{tabular}
\caption{CMU-MOSEI}
\label{tab:CMU-MOSEI7}
\end{subtable}
\vspace{0.3cm}
\begin{subtable}{\textwidth}
\centering
\small
\setlength{\tabcolsep}{3pt}
\renewcommand{\arraystretch}{1.2}
\begin{tabular}{lllllll}
\hline
 & $\mathbf{ACC_2}$ & $\mathbf{ACC_3}$ & $\mathbf{ACC_5}$ & $\mathbf{F1}$ & $\mathbf{MAE}$ & $\mathbf{Corr}$ \\ \hline
\textbf{Fused Features Only} & \textbf{82.93} & 69.37 & 49.38 & \textbf{82.9} & 33.2 & \textbf{73.26} \\ \hline
\textbf{Concatenation} & 82.42 & 69.44 & 49.82 & 82.38 & 33.6 & 72.87 \\ \hline
\textbf{Transformer} & 82.42 & \textbf{69.95} & \textbf{49.89} & 82.52 & \textbf{33.12} & 72.61 \\ \hline
\end{tabular}
\caption{CH-SIMS}
\label{tab:CH-SIMS5}
\end{subtable}
\caption{\textbf{Comparative Performance of Model Variations}: The {\it Fused Features Only} model employs only the features following the fusion network, while the {\it Concatenation} model merges the original signal with the fused signal. The {\it Transformer} model uses a transformer to combine these two signals. Across all metrics for both CMU-MOSEI and CH-SIMS, these three methods exhibit similar performance.}
\label{tab:Performance6}
\end{table*}

\begin{table*}
\centering

\begin{subtable}{\textwidth}
\centering
\small
\setlength{\tabcolsep}{3pt}
\renewcommand{\arraystretch}{1.2}
\begin{tabular}{lllllllll}
\hline
 \textbf{Context window} & $\mathbf{Has0\_ACC_2}$ & $\mathbf{Has0\_F1}$ & $\mathbf{Non0\_ACC_2}$ & $\mathbf{Non0\_F1}$ & $\mathbf{ACC_5}$ & $\mathbf{ACC_7}$ & $\mathbf{MAE}$ & $\mathbf{Corr}$ \\ \hline
\textbf{0} & 84.89 & 84.86 & 87.04 & 87.07 & \textbf{54.81} & \textbf{47.32} & 66.62 & 83.27 \\ \hline
\textbf{1} & \textbf{86.01} & \textbf{85.94} & \textbf{88.01} & \textbf{87.99} & 53.35 & 45.87 & \textbf{66.37} & \textbf{83.96} \\ \hline
\textbf{2} & 85.81 & 85.71 & 87.8 & 87.76 & 53.98 & 46.99 & 66.7 & 82.49 \\ \hline
\textbf{3} & 84.94 & 84.86 & 86.84 & 86.81 & 52.14 & 45.19 & 69.85 & 81.3 \\ \hline
\end{tabular}
\caption{Concatenation}
\label{tab:CMU-MOSEI5}
\end{subtable}
\vspace{0.3cm}

\begin{subtable}{\textwidth}
\centering
\small
\setlength{\tabcolsep}{3pt}
\renewcommand{\arraystretch}{1.2}
\begin{tabular}{lllllllll}
\hline
 \textbf{Context window} & $\mathbf{Has0\_ACC_2}$ & $\mathbf{Has0\_F1}$ & $\mathbf{Non0\_ACC_2}$ & $\mathbf{Non0\_F1}$ & $\mathbf{ACC_5}$ & $\mathbf{ACC_7}$ & $\mathbf{MAE}$ & $\mathbf{Corr}$ \\ \hline
\textbf{0} & 84.89 & 84.86 & 87.04 & 87.07 & 54.81 & 47.32 & 66.62 & 83.27 \\ \hline
\textbf{1} & 85.57 & 85.51 & 87.8 & 87.8 & \textbf{55.44} & \textbf{47.81} & 65.17 & 83.37 \\ \hline
\textbf{2} & \textbf{86.2} & \textbf{86.12} & \textbf{88.46} & \textbf{88.44} & 55.24 & 47.04 & \textbf{63.88} & \textbf{84.46} \\ \hline
\textbf{3} & 85.76 & 85.69 & 88.01 & 87.99 & 54.67 & 46.89 & 65.26 & 83.71 \\ \hline
\end{tabular}
\caption{Separation}
\label{tab:CMU-MOSEI6}
\end{subtable}

\caption{\textbf{Comparison of Context Modeling Methods with Only Text signals}: the second method that separates the context and the current utterance can handle a longer context window and have a better performance.}
\label{tab:Performance7}
\end{table*}

\begin{table*}
\centering
\small
\setlength{\tabcolsep}{3pt}
\renewcommand{\arraystretch}{1.2}

\begin{tabular}{lllllllll}
\hline
 \textbf{} & $\mathbf{Has0\_ACC_2}$ & $\mathbf{Has0\_F1}$ & $\mathbf{Non0\_ACC_2}$ & $\mathbf{Non0\_F1}$  & $\mathbf{ACC_7}$ & $\mathbf{MAE}$ & $\mathbf{Corr}$ \\ \hline
\textbf{fusion network} & \textbf{85.91} & \textbf{85.85} & \textbf{88.16} & \textbf{88.15}  & \textbf{48.25} & \textbf{64.29} & \textbf{83.8} \\ \hline
\textbf{- self attention layers} & 85.13 & 85.12 & 87.35 & 87.38 &46.94 & 65.81 & 81.53 \\ \hline
\textbf{- fully connected layers} & 85.13 & 85.09 & 87.2 & 87.2 & 46.65 & 66.48 & 83.07 \\ \hline
\end{tabular}

\caption{\textbf{Fusion Network Ablation Experiment Results}}
\end{table*}

\end{document}